\documentclass[runningheads,a4paper]{llncs}

\usepackage{amssymb}
\usepackage{amsmath}
\usepackage{graphicx}
\usepackage{caption}
\usepackage{subcaption}
\usepackage{multirow}
\usepackage{makecell}
\usepackage{verbatim}
\usepackage{hyperref}
\usepackage{gensymb}
\usepackage{mathtools}

\usepackage{color}

\DeclareMathOperator{\arctantwo}{arctan2}
\DeclareMathOperator{\Tr}{Tr}

\begin{document}

\mainmatter  

\title{Relative Camera Pose Estimation Using Convolutional Neural Networks}
\titlerunning{Relative Camera Pose Estimation Using Convolutional Neural Networks}  
%
\author{Iaroslav Melekhov\inst{1} \and Juha Ylioinas\inst{1} \and Juho Kannala\inst{1} \and Esa Rahtu\inst{2}}
\authorrunning{Relative Camera Pose Estimation Using Convolutional Neural Networks} 
%
%
\institute{Aalto University, Finland,\\
\email{firstname.lastname@aalto.fi}\\
\and
Tampere University of Technology, Finland\\
\email{esa.rahtu@tut.fi}}


%
%
%


%
%

\maketitle

\begin{abstract}
This paper presents a convolutional neural network based approach for estimating the relative pose between two cameras. The proposed network takes RGB images from both cameras as input and directly produces the relative rotation and translation as output. The system is trained in an end-to-end manner utilising transfer learning from a large scale classification dataset. The introduced approach is compared with widely used local feature based methods (SURF, ORB) and the results indicate a clear improvement over the baseline. In addition, a variant of the proposed architecture containing a spatial pyramid pooling (SPP) layer is evaluated and shown to further improve the performance.



\end{abstract}

\section{Introduction}

\begin{figure}[t!]
\centering
\includegraphics[width=.3\textwidth]{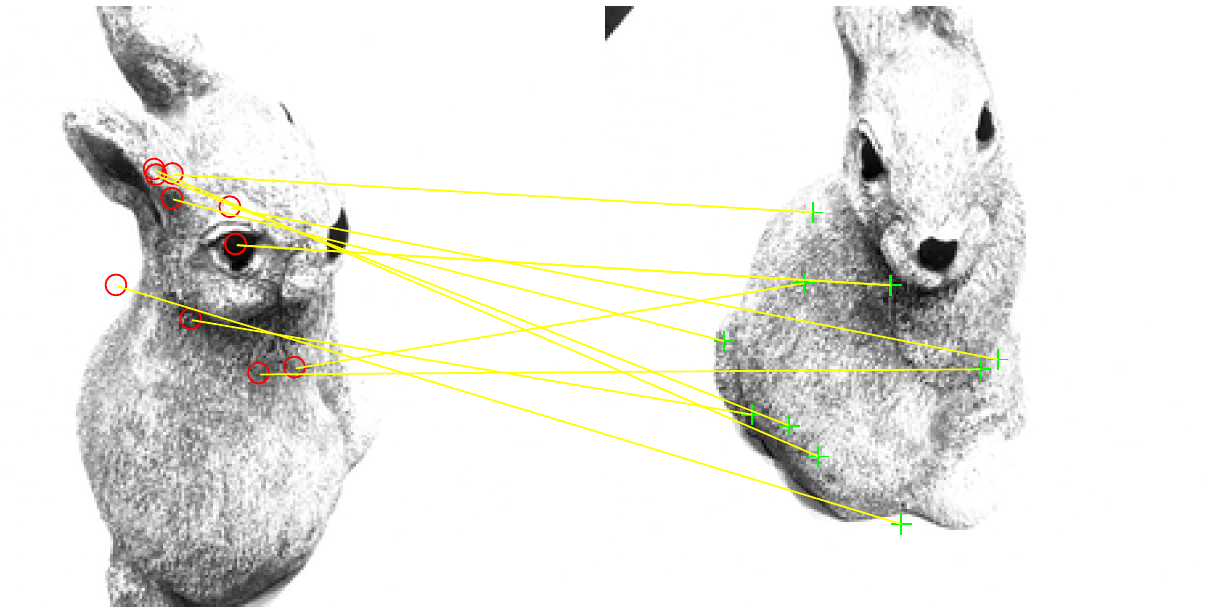}
\includegraphics[width=.3\textwidth]{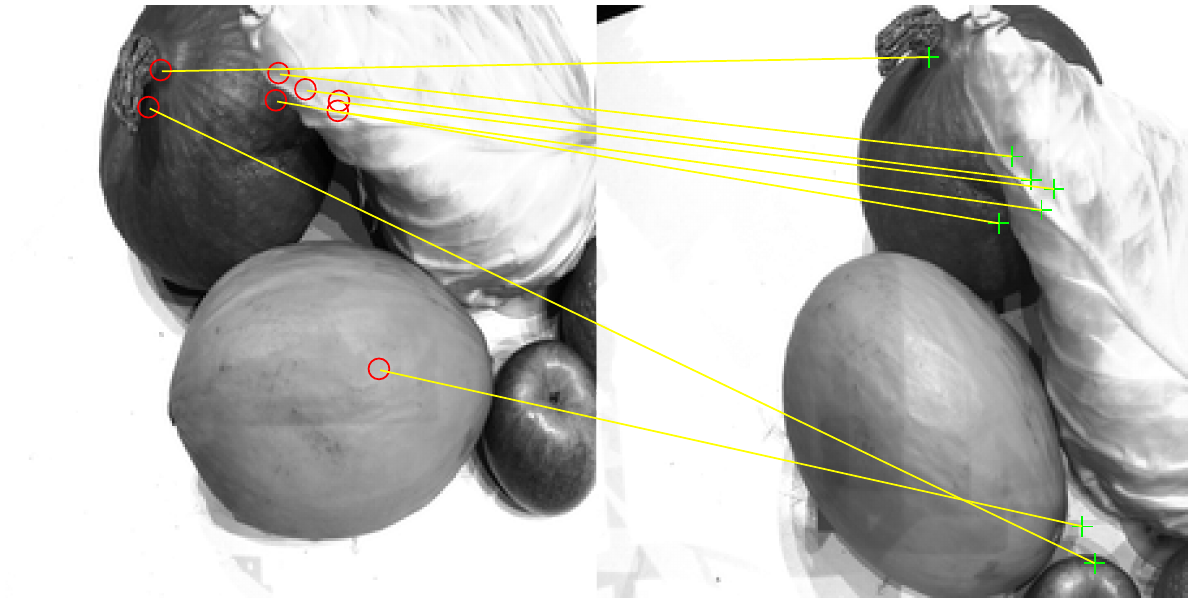}
\includegraphics[width=.3\textwidth]{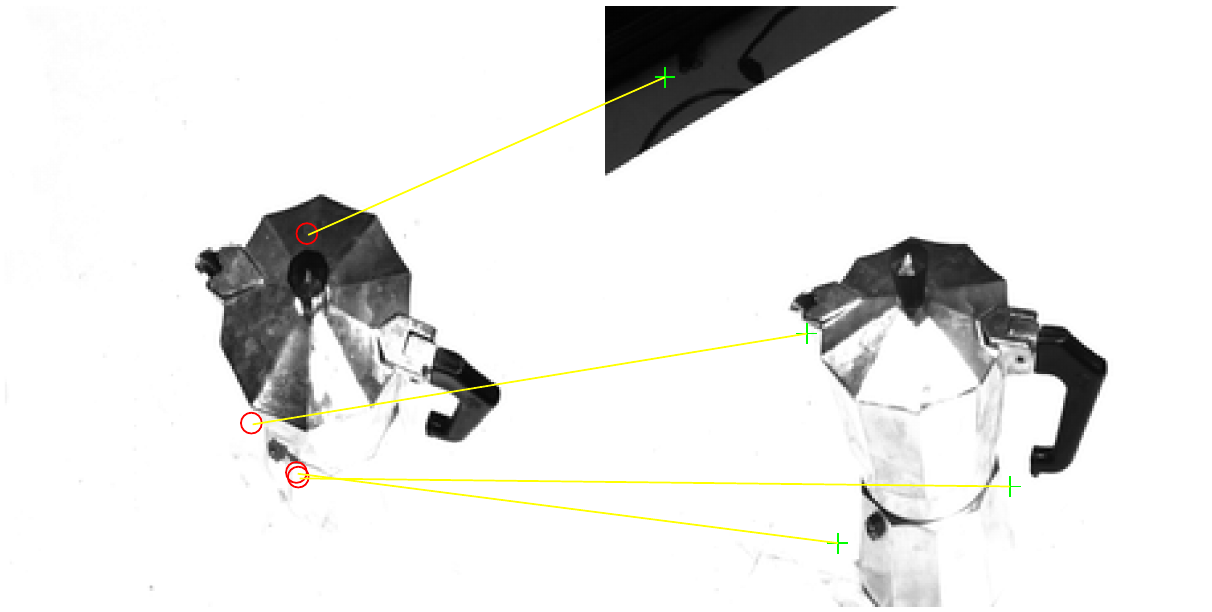}
\caption{Scenarios where traditional approaches are not able to estimate relative camera pose precisely. \textit{Left}: very large viewpoint changes, thus most of inliers (correspondences) are not correct; \textit{center}: the correct inliers concentrate on a small region; \textit{right}: there is insufficient number of correspondences due to textureless scene (object with reflecting surface).}\label{fig:fail_cases}
\end{figure}

The ability to estimate the relative pose between camera views is essential for many computer vision applications, such as structure from motion (SfM), simultaneous localization and mapping (SLAM) and visual odometry. Due to its practical importance, plenty of research effort has been devoted to the topic over the years. One popular approach to the problem is based on detecting and matching local feature points and using the obtained correspondences to determine the relative poses. The performance of such system is highly dependent on the accuracy of the local feature matches, which are commonly determined using methods like SIFT~\cite{SIFT}, DAISY~\cite{DAISY}, or SURF~\cite{SURF}. Unfortunately, there are many practically important cases where these hand-crafted descriptors are not able to find sufficient amount of correspondences. Particularly, repetitive structures, textureless objects, and extremely large viewpoint changes are difficult to handle. We highlight such cases in Fig.~\ref{fig:fail_cases}. An alternative solution would be to utilize all photometric information from the images to determine the poses. However, such methods (e.g.~\cite{visualSlam}) are usually not applicable to wide baseline settings, where there is large viewpoint change, or they may be computationally expensive. 

Recently, methods based on convolutional neural networks (CNNs) have clearly outperformed previous state-of-the-art results in many computer vision problems, such as image classification, object recognition, and image retrieval. In this work, we show how CNNs can also be applied to estimate the relative camera poses. Our contributions are as follows: 1) we propose a CNN-based method, which takes RGB images from both cameras as input and directly produces the relative rotation and translation as output; 2) we explore several network architectures and evaluate their performance on the DTU dataset~\cite{jensen2014large}; 3) we study how different training strategies affect the results and make comparisons to popular keypoint based approaches. 
 In addition, we investigate how spatial pyramid pooling~\cite{spp} could be applied in the context of relative camera pose estimation problem.

The rest of the paper is organized as follows. Section~\ref{sec:relatedWork} describes the related work focusing on relative camera pose estimation. The proposed approach and details related to network architectures and objective functions are introduced in Section \ref{sec:method}. Finally, sections \ref{sec:experiments} and \ref{sec:conc} present the baseline methods, experimental setup, evaluation results, discussion, and possible directions for future investigations.



\vspace{-2mm}
\section{Related Work}\label{sec:relatedWork}
\vspace{-2mm}
Over the years, a large variety of different local feature-based methods, such as SIFT \cite{SIFT}, SURF \cite{SURF}, ORB \cite{ORB}, BRIEF \cite{BRIEF}, have been utilized in structure from motion, image-based localization,  and visual SLAM contexts for estimating camera poses. The main disadvantage of these methods is their limited ability to cope with nuisance factors such as variations in viewpoint, reflections, and lack of distinguishable texture. As also noted in \cite{deepHomography}, some recent efforts indicate promise in approaching geometric computer vision tasks with a dense, featureless methods based on using full images. Taking this into account, one of the most prominent solutions is to apply convolutional neural networks (CNNs). While they have recently been applied in many computer vision problems, there are only a few works that consider them in the context of relative pose estimation. 

Konda et al.~\cite{konda} proposed a CNN architecture for predicting change in velocity and local change in orientation using short stereo video clips. They used a rather shallow CNN architecture together with unsupervised pre-training of the filters in early layers. Partly because of the shortage of training data in their path prediction application, they were forced to discretize the space of velocities and local changes for a softmax-based classification instead of continuous estimates with regression. Mohanty et al.~\cite{deepvo} tried to solve the same problem as in \cite{konda} using a monocular approach. In detail, they used an architecture based on two AlexNet-like CNN branches acting as inputs to a stack of fully connected layers coupled with a regression layer. 

Ummenhofer et al.~\cite{demon} proposed a CNN architecture for depth and relative camera motion estimation. They utilized multiple tasks in the learning phase to provide additional supervision in order to get more accurate depth maps and camera motion estimates. DeTone et al.~\cite{deepHomography} proposed a CNN architecture for estimating the relative homography between two images by regressing a 4-point homography parameterization with an Euclidean loss. Finally, instead of relative camera pose, Kendall et al.~\cite{kendall2015convolutional} proposed a CNN-based method for absolute 6-DoF camera pose estimation.

Our proposal is related to all previously discussed works, but it is the first one investigating the suitability of Siamese network architectures in the relative camera pose estimation problem. Compared with \cite{konda,deepvo}, our study aims at more general treatment of the camera pose estimation problem. That is, our approach is applicable for general unrestricted camera motion and for wide baseline view pairs, unlike \cite{konda,deepvo}. Compared with \cite{deepHomography}, we are trying to solve relative camera pose, which can be regarded as a more general problem than solving the relative homography between two views. Regarding \cite{kendall2015convolutional}, we adopt the same learning objective but concentrate on solving a different problem. In particular, unlike prediction of absolute pose \cite{kendall2015convolutional}, relative pose estimation provides means for relation and representation learning for \emph{previously unseen} scenes and objects. Finally, compared with \cite{demon}, our study focuses on analyzing the differences in traditional and CNN-based approaches for relative camera pose estimation and does not consider the role of additional supervisory signals. That is, our approach does not require depth maps for training which is beneficial in practice. Further details of our approach will be given in the following sections.

\section{Methodology}\label{sec:method}
%
Our goal is to estimate relative camera pose directly by processing a pair of images captured by two cameras. We propose a convolutional neural network based method that predicts a 7-dimensional relative camera pose vector $\Delta\mathbf{p}$ containing the relative orientation vector $\Delta\mathbf{q}$ (4-dimensional quaternion), and the relative position, i.e.\ translation vector $\Delta\mathbf{t}$ (3-dimensional), so that $\Delta\mathbf{p}=\left[\Delta\mathbf{q}, \Delta\mathbf{t}\right]$.

\subsection{Network Architectures}
\begin{figure}[t!]
\centering
\includegraphics[width=0.8\textwidth]{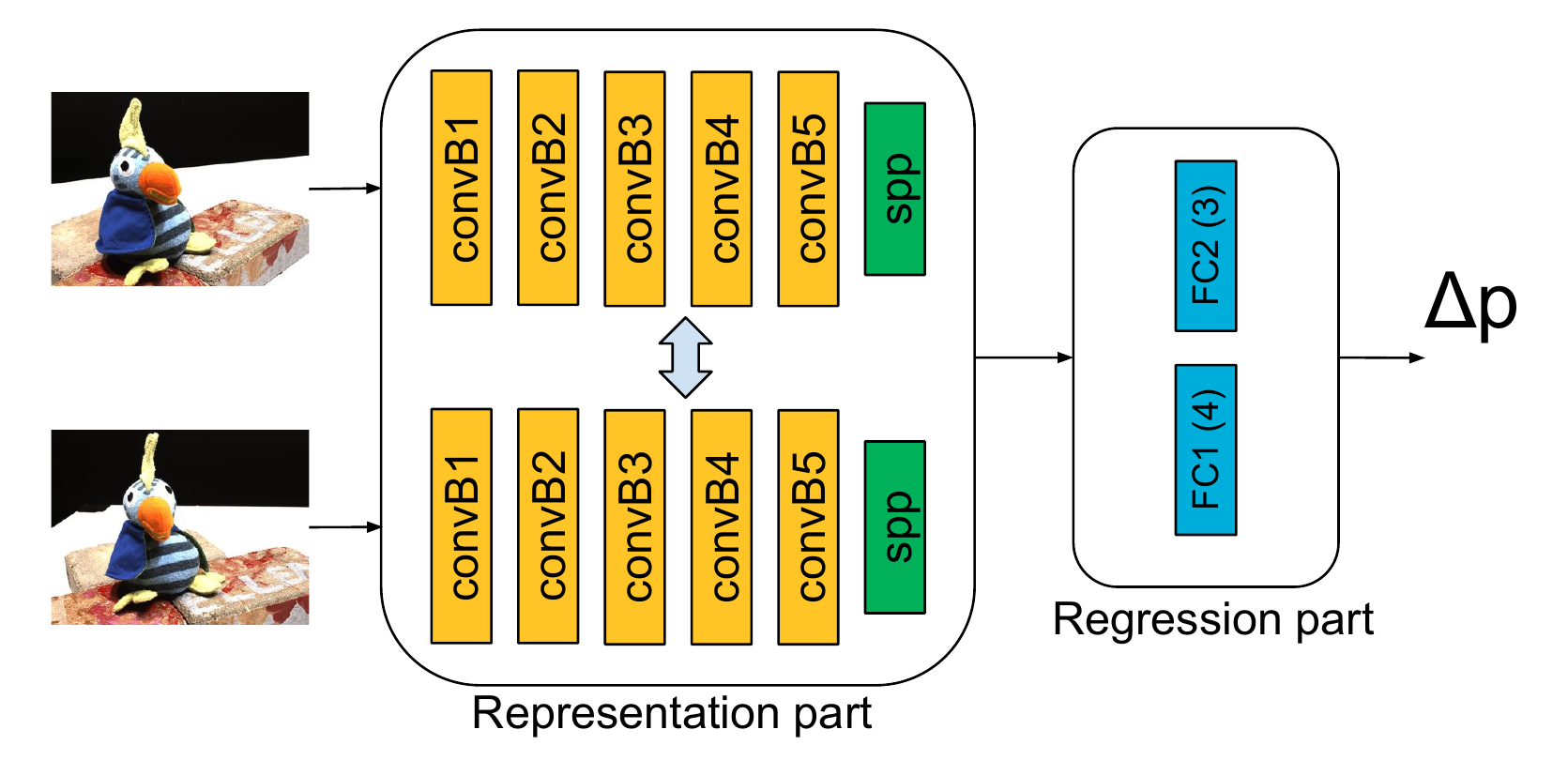}
\caption{Model structure (\textit{cnnBspp}). Both network branches (representation part) have identical structure with shared weights. Pre-trained Hybrid-CNN~\cite{hybridcnn} neural network was utilized to initialize the proposed architecture. Representation part maps an image pair to a low dimensional feature vector which is processed by regression part of the network. Regression part consists of 2 fully-connected layers (FC1 and FC2) and estimates relative camera pose.}\label{fig:network_architecture}.  
\end{figure}

To estimate the relative camera pose between a pair of images, we apply a Siamese network architecture~\cite{lecunSiameseNetwork} (see Fig.~\ref{fig:network_architecture}). In detail, our network consists of two blocks termed as the representation and the regression part. The representation part incorporates two identical CNN branches sharing the weights and other parameters. In general, both of these branches are composed of convolutional layers and rectified linear units (ReLU). The regression part in turn is composed of two fully-connected (FC1 and FC2) layers, where FC1 and FC2 have 4 and 3 connections respectively for estimating the 7-dimensional pose vector.

Following~\cite{kendall2015convolutional}, we apply transfer learning. In detail, we take the Hybrid-CNN~\cite{hybridcnn} as a base network for both of the branches and initialize them based on weights learned usin large-scale classification data. More specifically, Hybrid-CNN is AlexNet trained on both image classification \textit{ImageNet}~\cite{Krizhevsky} and a scene-centric \textit{Places}~\cite{hybridcnn} datasets.

Extracting features from convolutional layers instead of fully-connected layers has shown to produce more accurate results in image retrieval problem~\cite{babenko,razavian,azizpour}. Therefore, we removed the last three fully-connected layers (FC6, FC7 and FC8) from the original Hybrid-CNN preserving only convolutional, max-pooling layers and ReLU. More precisely, the network architecture for one branch has the following blocks: convB1$_{[96,11,4,0]}$, pool$_{[3,2]}$, convB2$_{[256,5,1,2]}$, pool$_{[3,2]}$, convB3$_{[384,3,1,1]}$, convB4$_{[384,3,1,1]}$, convB5$_{[256,3,1,1]}$, pool$_{[3,2]}$. The notation convB$_{[N,\omega,s,p]}$ consists of a convolution layer with $N$ filters of size $\omega\times\omega$ with stride $s$ and padding $p$ and a regularization layer (ReLU), pool$_{[k,s]}$ is a max-pooling layer of size $k\times k$ applied with a stride $s$. The last layer of this baseline architecture dubbed \textit{cnnA} is a pooling layer producing a tiny feature map $(6\times6)$ as an output. Therefore, due to reduction of spatial resolution, image details that are beneficial for relative camera pose estimation may have been lost at this part of the network. In order to limit such information loss, we remove the last max-pooling layer extracting features from convB5, which allows to have slightly larger feature maps (size $13\times13$). This modified version of the \textit{cnnA} architecture is called \textit{cnnB}.

Each branch of our representation part has an AlexNet-like structure originally designed for a fixed-size $(227\times227)$ input image. Such limitation may lead to reduced accuracy in the relative camera pose estimation degrading the performance of the system in general. To have more accurate estimations, it might be beneficial to process larger images to be able to extract more information from the scene structure. Theoretically, the convolutional layers accept arbitrary input image sizes, but they also produce outputs of variable dimensions. However, the FC layer of the regression part (see Fig.~\ref{fig:network_architecture}) requires to have fixed-length vectors as input. To make our pipeline to accept arbitrary image sizes, we apply a spatial pyramid pooling (SPP) layer which can maintain spatial information by pooling in local spatial bins~\cite{spp}. An SPP layer consists of a set of pooling layers of $n\times n$ bins with the window size $\mathit{w}=ceil(a/n)$ and a stride $\mathit{str}=floor(a/n)$, where $a$ is a size of the input feature map $\left(a\times a\right)$ of the SPP layer. Therefore, the number of output bins of the SPP layer is fixed regardless of the image size.

We modified the original architectures \textit{cnnA} and \textit{cnnB} by adding an SPP layer to the end of each branch. Obtained networks (\textit{cnnAspp} and \textit{cnnBspp}) have 4-level $\left(1\times1,\, 2\times2,\, 3\times3,\, 6\times6\right)$ and 5-level \\$\left(1\times1,\, 2\times2,\, 3\times3,\, 6\times6,\, 13\times13\right)$ pyramid pooling respectively. An \textit{cnnBspp} structure is illustrated in Fig.~\ref{fig:network_architecture}. More detailed evaluation of the proposed network architectures is presented in Sec.~\ref{ssec:dl_models_comparison}.


\subsection{Learning and inference}
To regress the relative pose, the network was designed to compute the Euclidean loss between estimated vectors and ground truth. Following~\cite{kendall2015convolutional}, we predict the relative orientation and position together using only one neural network learnt in a multi-task fashion. 

Our loss function for training is as follows
\begin{equation}\label{eq:loss_function}
\mathcal{L} = \left\|\Delta\mathbf{\hat{t}} - \Delta\mathbf{t}\right\|_2 + \beta\left\|\Delta\mathbf{\hat{q}} - \Delta\mathbf{q}\right\|_2
\end{equation}
where $\Delta\mathbf{\hat{q}}$ and $\Delta\mathbf{\hat{t}}$ are the ground-truth relative orientation and translation, and $\beta$ is a parameter to keep the estimated values to be nearly equal. As described in~\cite{kendall2015convolutional}, $\beta$ must balance the orientation and translation estimations and can be set by grid search. In our experiments we set $\beta$ equal to 10. The network is trained via back-propagation using stochastic gradient descent. The details of the training are described in Section \ref{ssec:dl_models_comparison}.

It should be noted that quaternion vectors have unit length by definition and therefore $||\Delta\mathbf{\hat{q}}||\!=\!1$ in \eqref{eq:loss_function}. Further, since the absolute scale of the translation can not be recovered, we normalize the ground truth translations to unit length as well, i.e.\ $||\Delta\mathbf{\hat{t}}||\!=\!1$. However, the norm constraints are not explicitly enforced during training. Instead, the estimates are normalized to unit length as a post-processing step like in \cite{kendall2015convolutional}.

Thus, at test time, a pair of images is fed to a regression neural network, consisting of two branches, which directly estimates the real-valued parameters of the relative camera pose vector. Finally, the estimated quaternion and translation vectors are normalized to unit length.

\subsection{Error metrics}

The error metrics that we use for evaluation of the performance are: (a) relative orientation error (ROE) in degrees and (b) relative translation error (RTE) in degrees. The latter error is the angle between the estimated translation vector and the ground truth. The former error is measured by determining the rotation angle for the rotation between the estimated orientation and the ground truth (i.e.\ 
we determine the rotation which rotates the estimated orientation onto the ground truth).

\begin{figure}[t!]
\centering
\begin{subfigure}[]{.32\textwidth}
\centering
\includegraphics[width=.5\textwidth]{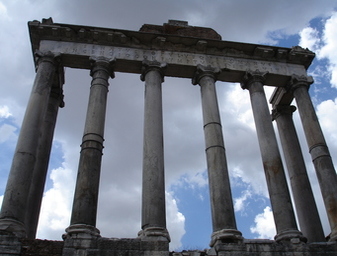}%
\includegraphics[width=.5\textwidth]{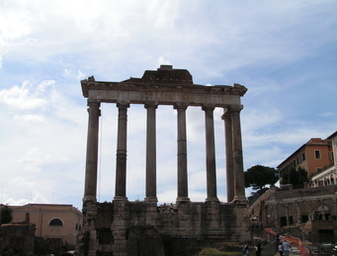}
\caption{Roman Forum}\label{subfig:train_examples_1}
\end{subfigure}
\begin{subfigure}[]{.32\textwidth}
\centering
\includegraphics[width=.5\textwidth]{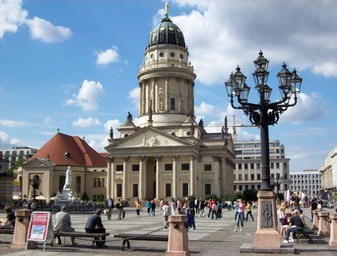}%
\includegraphics[width=.5\textwidth]{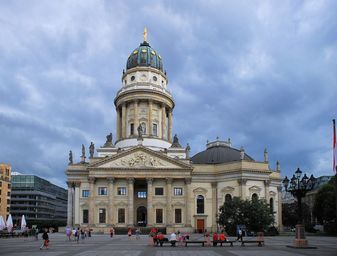}
\caption{Gendarmenmarkt}\label{subfig:train_examples_2}
\end{subfigure}
\begin{subfigure}[]{.32\textwidth}
\centering
\includegraphics[width=.5\textwidth]{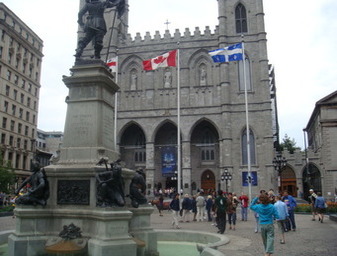}%
\includegraphics[width=.5\textwidth]{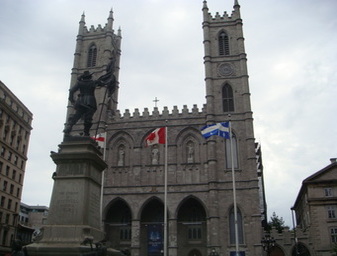}
\caption{Montreal Notre Dame}\label{subfig:train_examples_3}
\end{subfigure}
\begin{subfigure}[]{.32\textwidth}
\centering
\includegraphics[width=.5\textwidth]{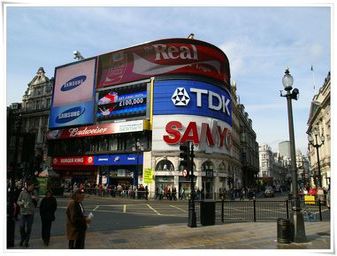}%
\includegraphics[width=.5\textwidth]{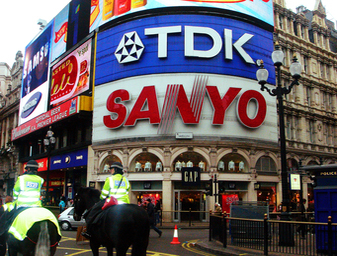}
\caption{Piccadilly}\label{subfig:train_examples_4}
\end{subfigure}
\begin{subfigure}[]{.32\textwidth}
\centering
\includegraphics[width=.5\textwidth]{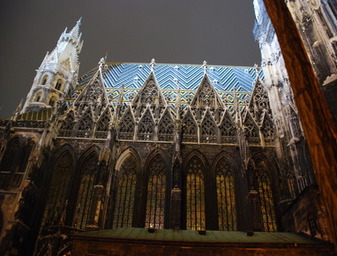}%
\includegraphics[width=.5\textwidth]{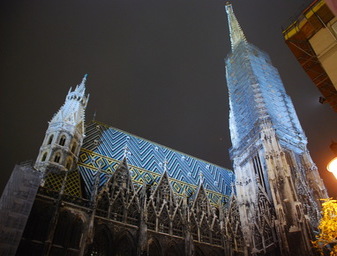}
\caption{Vienna Cathedral}\label{subfig:train_examples_5}
\end{subfigure}
\begin{subfigure}[]{.32\textwidth}
\centering
\includegraphics[width=.35\textwidth]{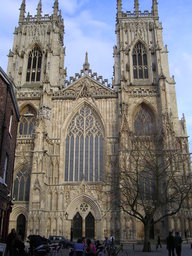}%
\includegraphics[width=.35\textwidth]{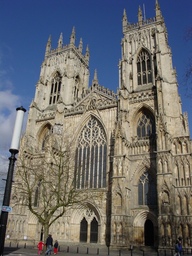}
\caption{Yorkminster}\label{subfig:validation}
\end{subfigure}
\caption{Examples of the training (\ref{subfig:train_examples_1}-\ref{subfig:train_examples_5}) and the validation (\ref{subfig:validation}) sets representing image pairs of six landmarks. The images were taken under different lighting and weather conditions, with variations of appearance and camera positions. Additionally, the dataset has a lot of occluded image pairs, so the problem of estimation relative camera pose becomes more challenging.}\label{fig:training_dataset}
\end{figure}

\subsection{Datasets}\label{ssec:datasets}
It is essential to have a large and consistent dataset for training CNNs for relative camera pose problem. However, collecting such data may be expensive and laborious task. 
We overcome this by utilizing a crowd-sourced image collection provided by~\cite{1dsfm}, where the ground truth camera poses are obtained by using an automatic structure from motion pipeline based on local feature matching. The collection consists of 13 subsets of images representing different landmarks and the numerical data of the global structure from motion reconstruction for each subset. 

To evaluate the proposed CNN architectures we construct datasets for training and validation. The training set is composed of samples of five landmarks (Montreal Notre Dame, Piccadilly, Roman Forum, Vienna Cathedral and Gendarmenmarkt) covering altogether 581k image pairs (see examples of each landmark in Fig.~\ref{fig:training_dataset}). For the validation set, we used the Yorkminster subset covering 22k image pairs in total. The ground truth labels (relative orientation and translation) are provided by \cite{1dsfm} and were computed by applying the SIFT keypoint detector and descriptor followed by the structure and motion estimation via 
RANSAC, triangulation and iterative optimization.

\begin{figure}[t!]
\centering
\includegraphics[width=.19\textwidth]{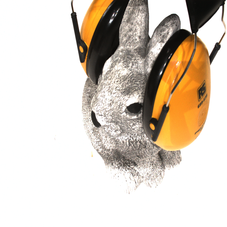}
\includegraphics[width=.19\textwidth]{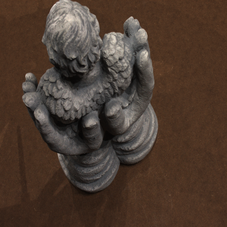}
\includegraphics[width=.19\textwidth]{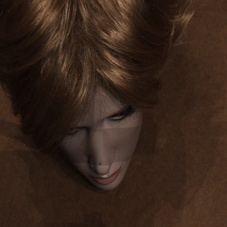}
\includegraphics[width=.19\textwidth]{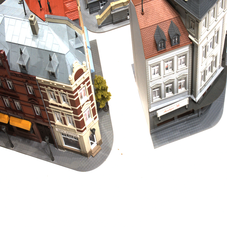}
\includegraphics[width=.19\textwidth]{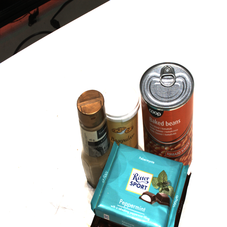}
\caption{Example scenes from the DTU Robot Image Dataset~\cite{jensen2014large}. The images show different objects which have been used in the evaluation dataset to estimate relative camera poses. In the dataset, camera positions are estimated very accurately as the camera was mounted on an industrial robot.}\label{fig:evaluation_dataset_examples}
\end{figure}

In order to obtain a fair comparison between our approach and point-based methods, we need to specify an evaluation dataset where the ground truth orientation and translation are accurate and reliable, and not dependent on the success of traditional point-based motion estimation methods.  
 For this, we utilize the DTU Robot Image Dataset provided by \cite{jensen2014large}, where the ground truth is obtained by attaching the camera to a precisely positioned robot arm. The dataset covers 124 scenes containing different number of camera positions. Several object scenes of this dataset are illustrated in Fig.~\ref{fig:evaluation_dataset_examples}. See \cite{jensen2014large} for further details about the pipeline that was used to collect the DTU dataset.

\vspace{-2mm}
\section{Experiments}\label{sec:experiments}
\vspace{-2mm}
We evaluated the performance of our proposal on the DTU dataset comparing it with two traditional feature based methods, namely SURF~\cite{SURF} and ORB~\cite{ORB}.

\vspace{-2mm}
\subsection{Preprocessing of the Evaluation Dataset} 
\vspace{-2mm}
As explained, the DTU dataset consists of 124 scenes covering different number of camera positions. More specifically, it contains 77 scenes (type-I) with 49 camera positions and 47 scenes (type-II) with 64 camera positions. In order to estimate relative camera poses between pairs of views, we first determine the camera pairs which have overlapping fields of view.


Assuming that the pairwise relationships between cameras are represented in a symmetric $n\times n$ adjacency matrix, it is easy to see that the upper bound for the number of overlapping view pairs is $n\left(n-1\right)/2$, where $n$ is the number of camera positions in the scene (49 or 64). Depending on the scene, this equals to 1176 and 2016, respectively. However, we compute relative poses only for the subset of pairs which have overlapping fields of view.  
 These camera pairs can be determined easily since the intrinsic camera parameters are known for the dataset. In detail, we define the field of view of each camera as a cone, i.e.\ a frustum, and calculate the intersection for pairs of corresponding frustums by using the publicly available OpenMVG~\cite{openMVG} library. As a result, the number of overlapping pairs of views is 512 for scenes of type-I and 753 for type-II. 

\vspace{-2mm}
\begin{figure}[t!]
\centering
\includegraphics[width=.45\textwidth]{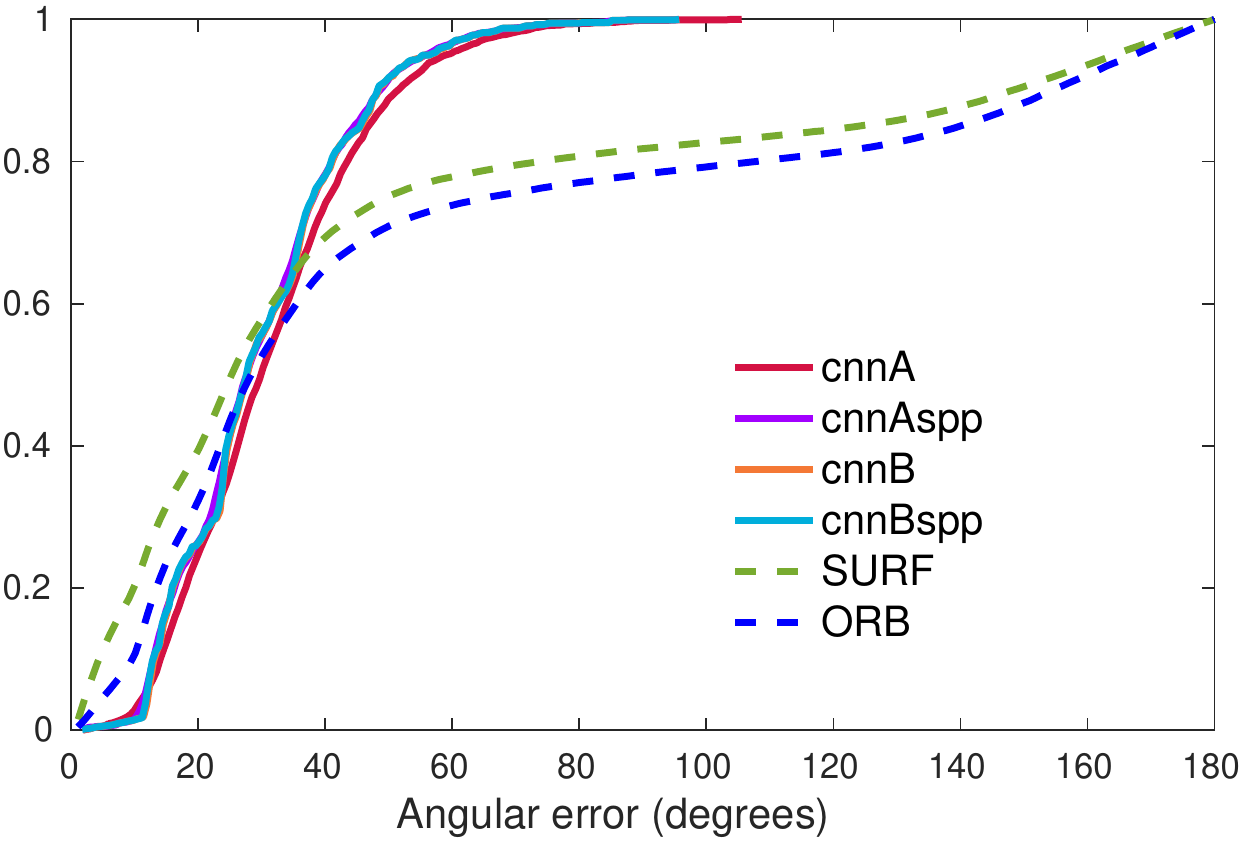}
\includegraphics[width=.45\textwidth]{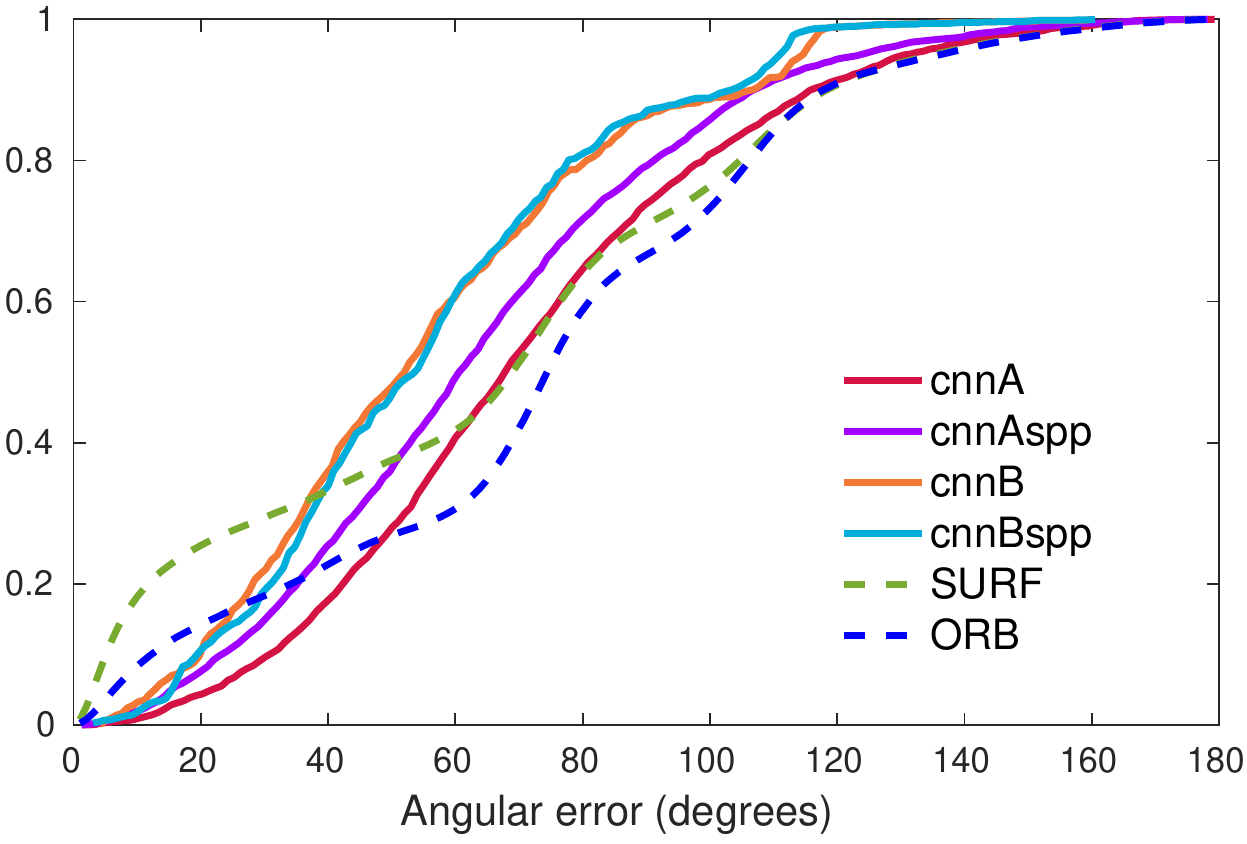}
\caption{Accuracy of our Siamese network architectures for estimating the relative camera orientation (left) and translation (right). Presented is the normalized cumulative histograms of errors for all scenes of the DTU dataset.}\label{fig:dl_models_comparison}
\end{figure}

\subsection{Comparing CNN models}\label{ssec:dl_models_comparison}
To compare the discussed Siamese network architectures with and without SPP, i.e. cnnA/cnnB and cnnAspp/cnnBspp (see Sec.~\ref{sec:method}), we created training image pairs by rescaling the input images so that the smaller dimension was fixed to 323 pixels and keeping the original aspect ratio. Depending on the model, we then used either random $227\times227$ or $323\times323$ pixel crops for training (i.e.\ the larger size for architectures applying SPP), and a center crop $\left(227\times227\right)$ at the test time. To train our networks we used stochastic gradient descent (SGD) and the Adam solver~\cite{Adam}. Learning rate $\left(10^{-4}\right)$, weight decay $\left(10^{-5}\right)$ as well as the size of mini-batches $\left(128\right)$ were the same for all models during the training phase. We used the publicly available machine learning framework Torch~\cite{Torch7} on two NVIDIA Titan X cards applying the data parallelism trick to speed up training. It took around 60 hours to finish 15 training epochs with the 581k image pairs of the training dataset described in Section \ref{ssec:datasets}.

Fig.~\ref{fig:dl_models_comparison} shows a set of normalized cumulative histograms of relative orientation and translation errors for each of the discussed models evaluated on all scenes of the DTU dataset. According to the results it can be seen that having a bigger output feature map size before the final FC layers is clearly beneficial. This can be seen especially in the case of the reported error on relative translations (Fig.~\ref{fig:dl_models_comparison} right) where the cnnB and cnnBspp outperform cnnA and cnnAspp. In addition, utilizing SPP yields a further improvement. 
 We dubbed the top-performing model (cnnBspp) to \texttt{cnn-spp} and used it for further experiments reported in the following section.

\begin{figure}[t!]
\centering
\begin{subfigure}[]{.9\textwidth}
\centering
\includegraphics[width=.4\textwidth]{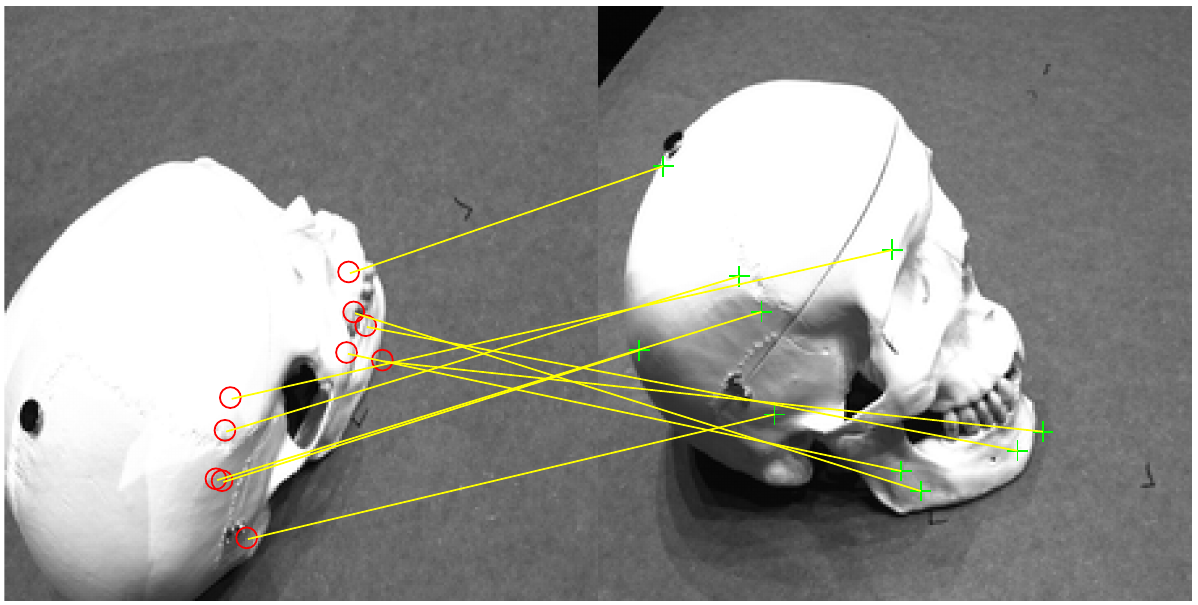}
\includegraphics[width=.4\textwidth]{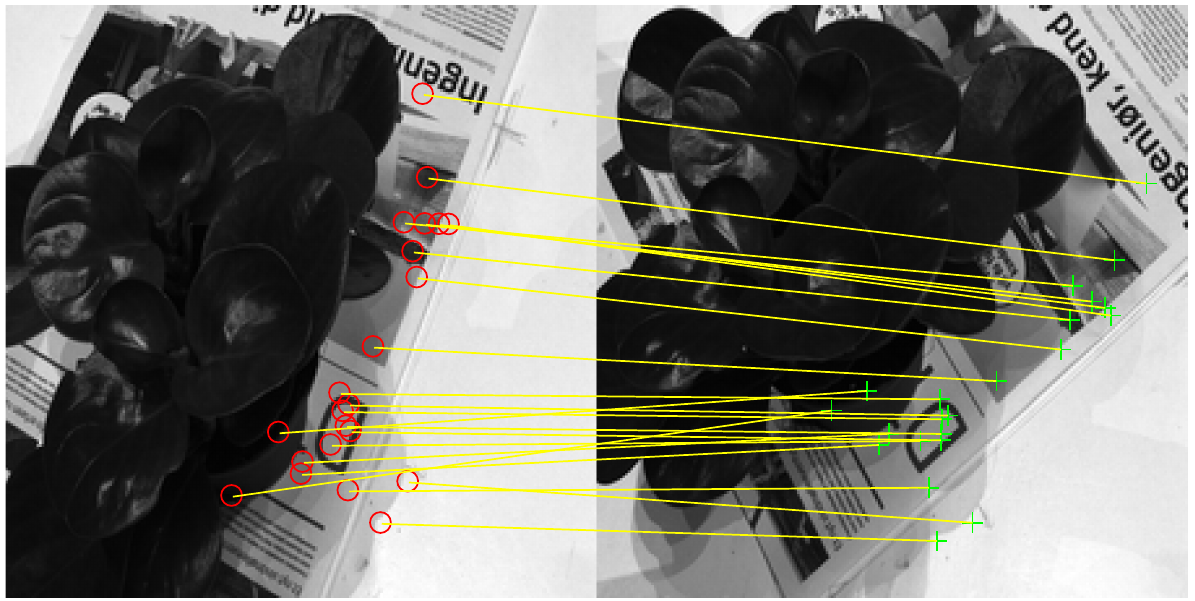}
\caption{Failure cases for which the traditional SURF approach was not able to detect enough matching points (inliers) properly or they were not distributed well in the image pair. As a result, the method has poor performance relative to the proposed method. ROE: $92.35^\circ\left(\mathit{52.66}^\circ\right)$; $57.86^\circ\left(\mathit{29.27}^\circ\right)$. RTE: $113.62^\circ\left(\mathit{34.49}^\circ\right)$; $118.9^\circ\left(\mathit{56.42}^\circ\right)$}\label{subfig:pb_fails}
\end{subfigure}
\begin{subfigure}[]{.9\textwidth}
\centering
\includegraphics[width=.4\textwidth]{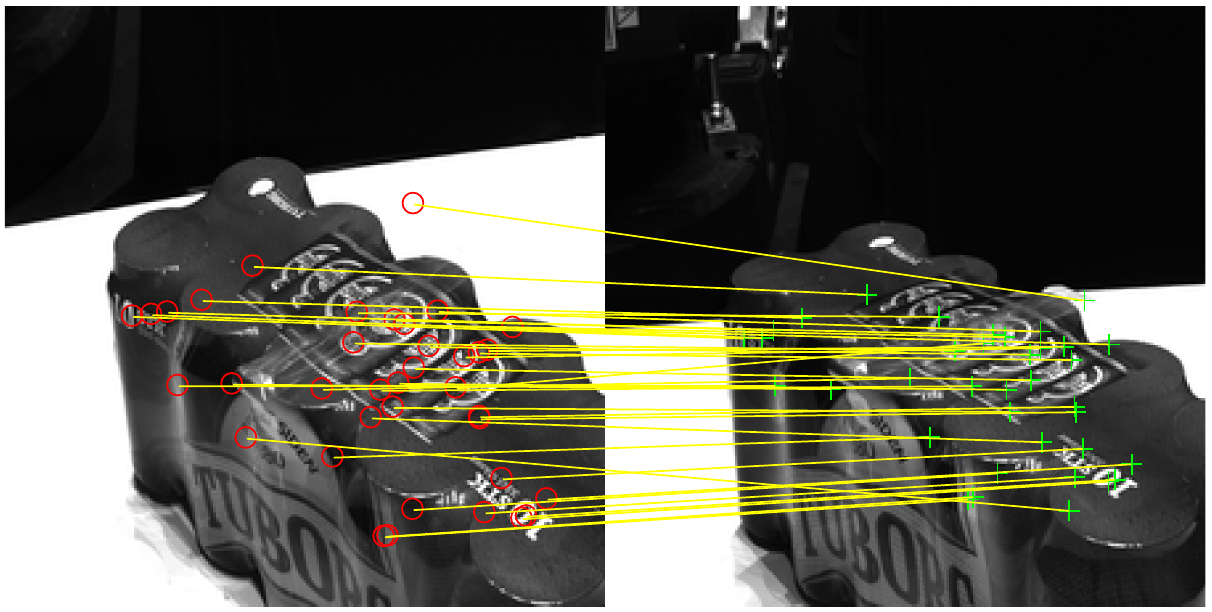}
\includegraphics[width=.4\textwidth]{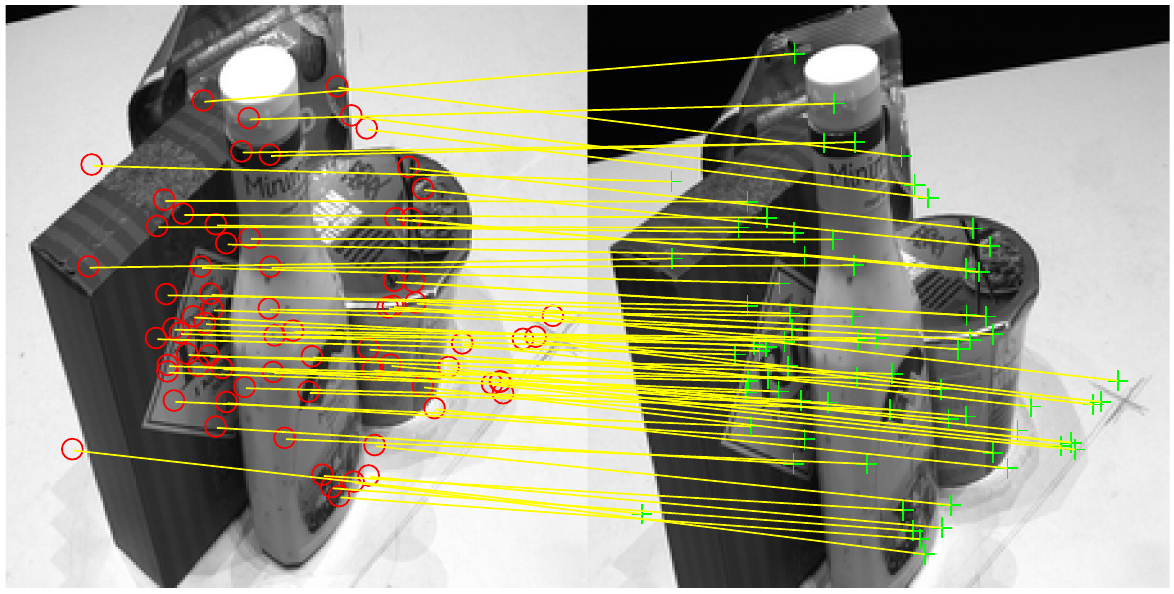}
\caption{Both approaches produce competitive results. ROE: $9.05^\circ\left(\mathit{12.36}^\circ\right)$; $19.93^\circ\left(\mathit{12.9}^\circ\right)$. RTE: $13.51^\circ\left(\mathit{15.10}^\circ\right)$; $53.74^\circ\left(\mathit{58.09}^\circ\right)$}\label{subfig:pb_equals}
\end{subfigure}
\begin{subfigure}[]{.9\textwidth}
\centering
\includegraphics[width=.4\textwidth]{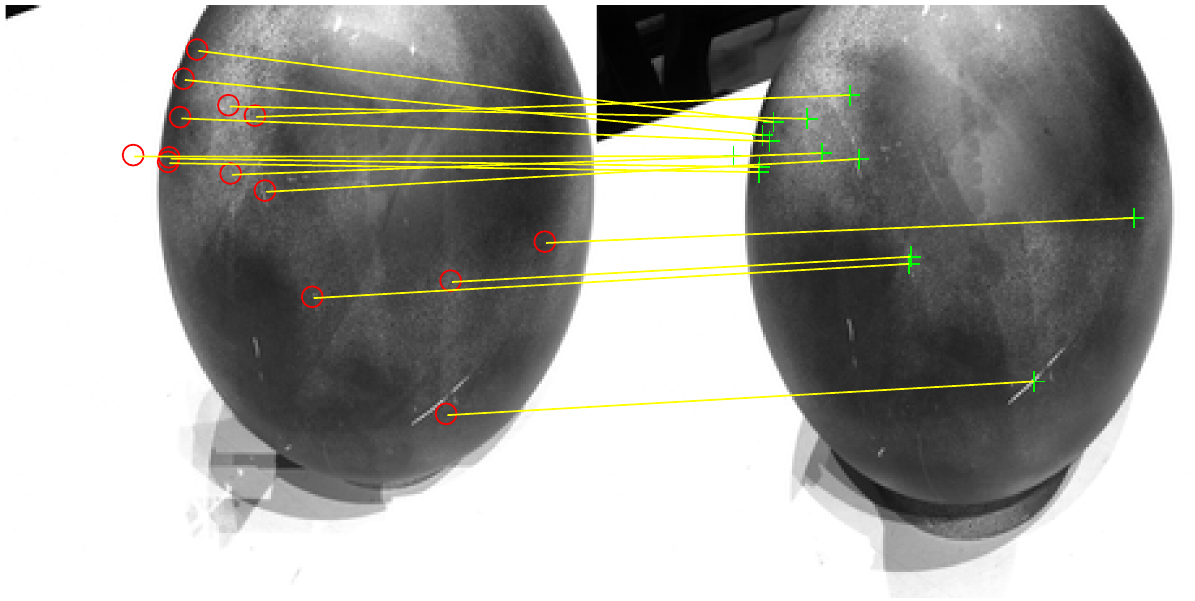}
\includegraphics[width=.4\textwidth]{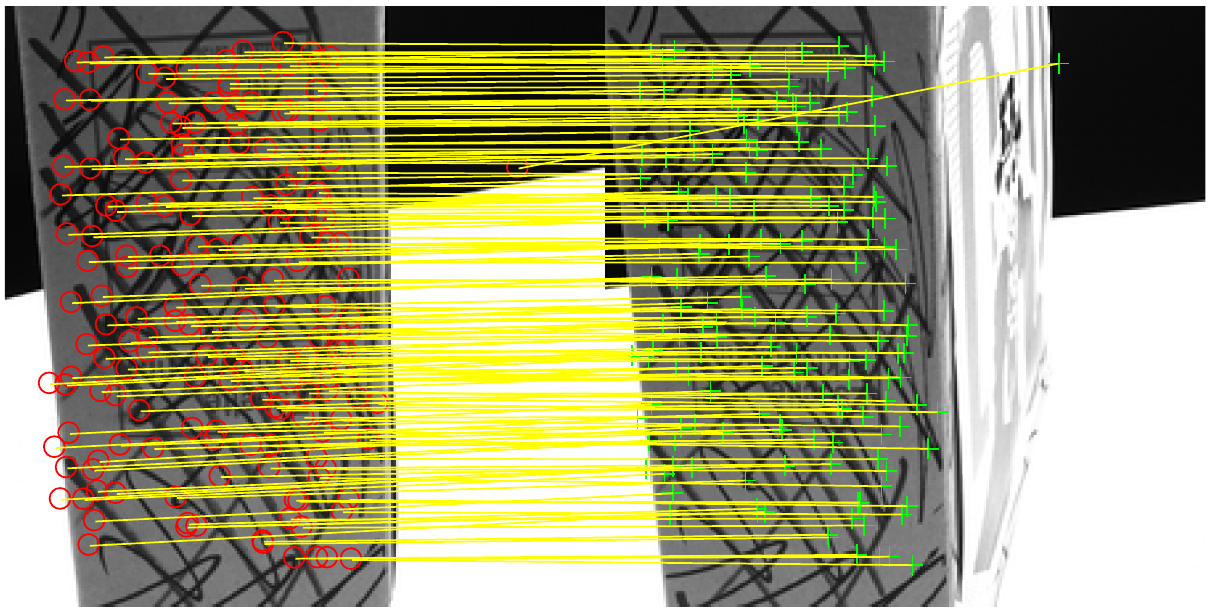}
\caption{Point-based descriptor finds sufficient amount of well-distributed features and outperforms our approach in relative translation estimation. ROE: $10.53^\circ\left(\mathit{12.52}^\circ\right)$; $9.24^\circ\left(\mathit{11.55}^\circ\right)$. RTE: $3.23^\circ\left(\mathit{41.15}^\circ\right)$; $15.42^\circ\left(\mathit{57.62}^\circ\right)$}\label{subfig:pb_outperforms}
\end{subfigure}
\caption{Visual illustration of the performance of a baseline (SURF) on example image pairs. For all the cases (\ref{subfig:pb_fails}, \ref{subfig:pb_equals}, \ref{subfig:pb_outperforms}) we also report the error measures (ROE and RTE) of the baseline and our best model (in \textit{parentheses}) on the given pairs.}\label{fig:bad_neutral_good_cases}
\vspace{-3mm}
\end{figure}

\vspace{-2mm}
\subsection{Comparison to traditional methods}
\vspace{-2mm}
We compare our best model with a baseline consisting of two feature based methods, namely SURF and ORB. For both of these methods, we used the OpenCV implementation with the default parameters. The pose was recovered from the essential matrix estimated using the five-point method and RANSAC.

The results presented in Fig.~\ref{fig:dl_models_comparison} confirm that our proposed model \texttt{cnn-spp} performs better compared with the baseline feature based methods. For fair comparison, we resize all images of the DTU dataset to $227\times227$ size, transform internal camera parameters accordingly, and evaluate CNNs and feature-based approaches on this data. Our results show that transfer learning from external datasets can be utilized effectively to train a general relative pose regressor. Further, it should be noted that essential matrix estimation requires knowledge of the internal camera parameters, whereas our CNN based approach does not use that information.

We illustrate example results in Fig.~\ref{fig:bad_neutral_good_cases}. The yellow lines show matching points of SURF features across the images in a pair. The visualization shows that our method is robust and in some cases produces more accurate relative pose estimations than conventional point-based methods (Fig.~\ref{subfig:pb_fails}).

In Sec.~\ref{ssec:datasets} we described the data used to train our models in the previous experiments. However, the visual characteristics and distribution of relative poses in this data are quite different from the data used in evaluation, namely the DTU dataset. 
Therefore, given that many studies~\cite{kendall2015convolutional,Krizhevsky} show that it is essential to use training data which is as close as possible to the data in the target domain, it is interesting to see how the performance is affected if a subset of DTU scenes is used for training. Thus, we  divided the DTU scenes into two sets, and used one of them to fine-tune cnn-spp model pre-trained on Landmarks dataset (Sec.~\ref{ssec:datasets}). 
 The other part of the DTU images was then used as the final testing set. Furthermore, according to the properties of SPP layer we conduct experiments using different input image sizes for the proposed model. Particularly, low resolution images $\left(227\times227\right)$ and high resolution $\left(1600\times1200\right)$ images were evaluated. The  size of high resolution images corresponds to the original images in the test set of the DTU dataset.

\begin{figure}
\centering
\begin{subfigure}[]{.9\textwidth}
\centering
\includegraphics[width=.40\textwidth]{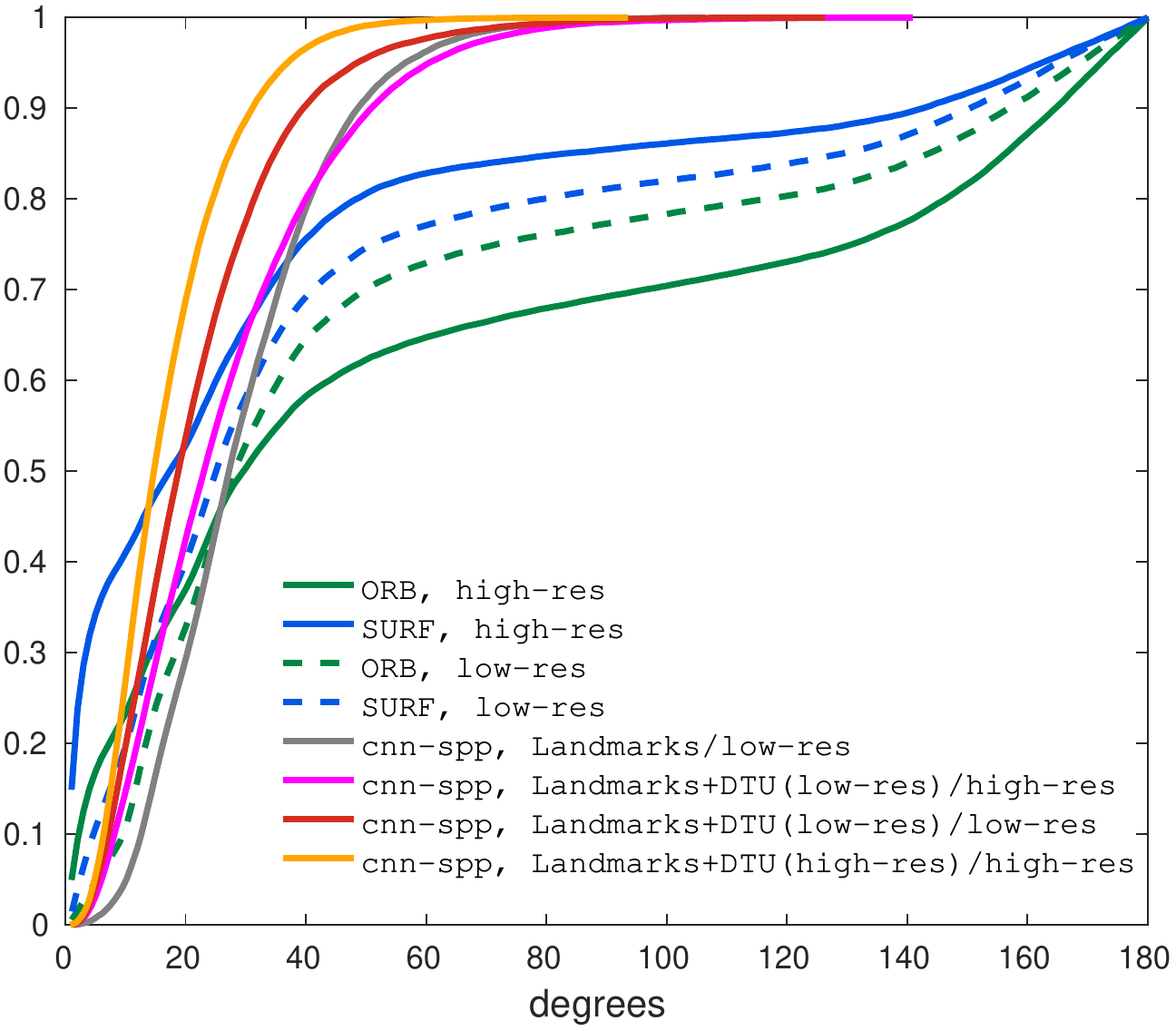}
\includegraphics[width=.55\textwidth]{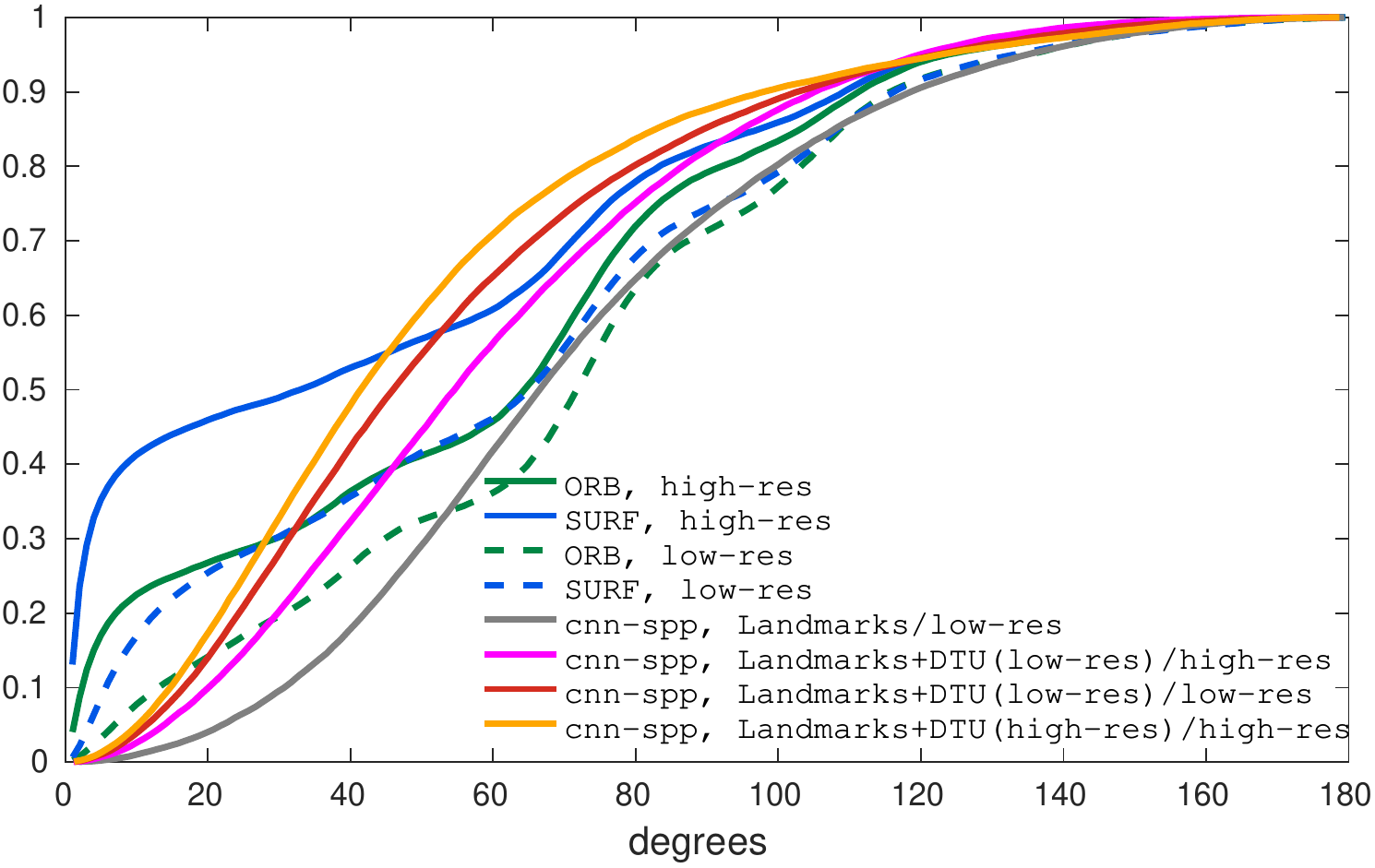}
\caption{Cumulative histogram of errors for relative camera orientation (left) and relative translation (right) for the test scenes of the DTU dataset. Experiments were evaluated on two different pre-defined image sizes: $227\times227$ (low-res) and $1600\times1200$ (high-res). The notation for CNN approaches is following: model, training data and the training image size (high-res or low-res), and the size of test images.}\label{subfig:performance_all_test_scenes}
\end{subfigure}
\begin{subfigure}[]{.9\textwidth}
\centering
\includegraphics[width=.25\textwidth]{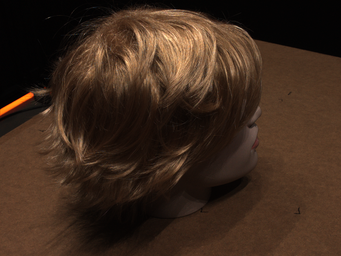}
\includegraphics[width=.25\textwidth]{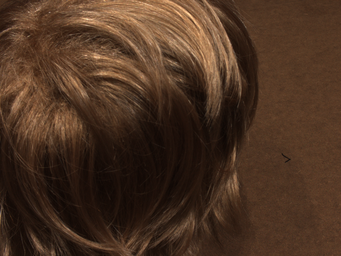}
\includegraphics[width=.25\textwidth]{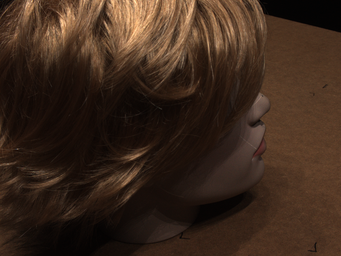}
\includegraphics[width=.25\textwidth]{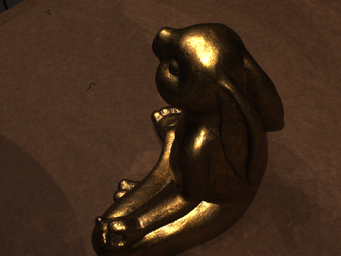}
\includegraphics[width=.25\textwidth]{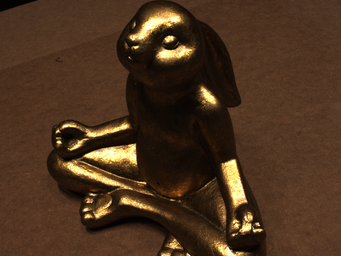}
\includegraphics[width=.25\textwidth]{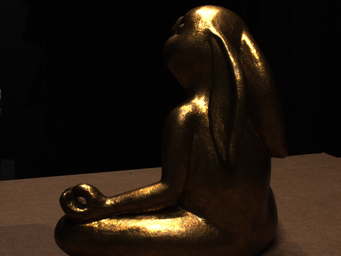}
\caption{Some scenes from the test set of the DTU dataset representing textureless objects with light reflections.}\label{subfig:head_exmpl}
\end{subfigure}
\begin{subfigure}[]{.9\textwidth}
\centering
\includegraphics[width=.45\textwidth]{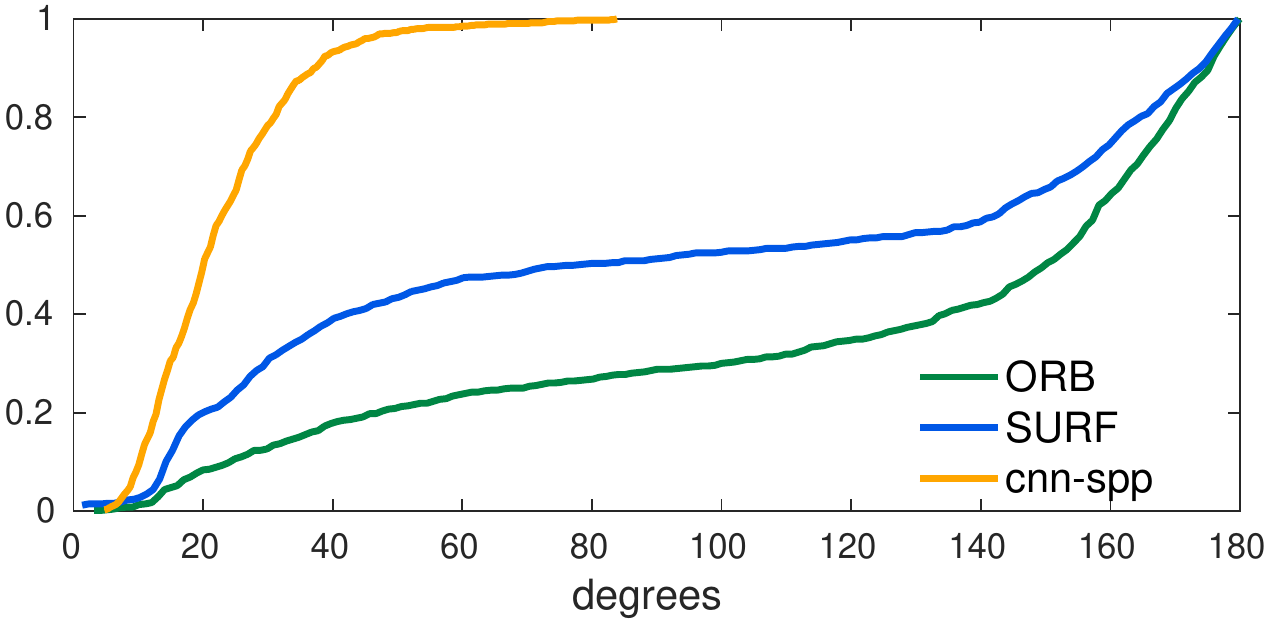}
\includegraphics[width=.45\textwidth]{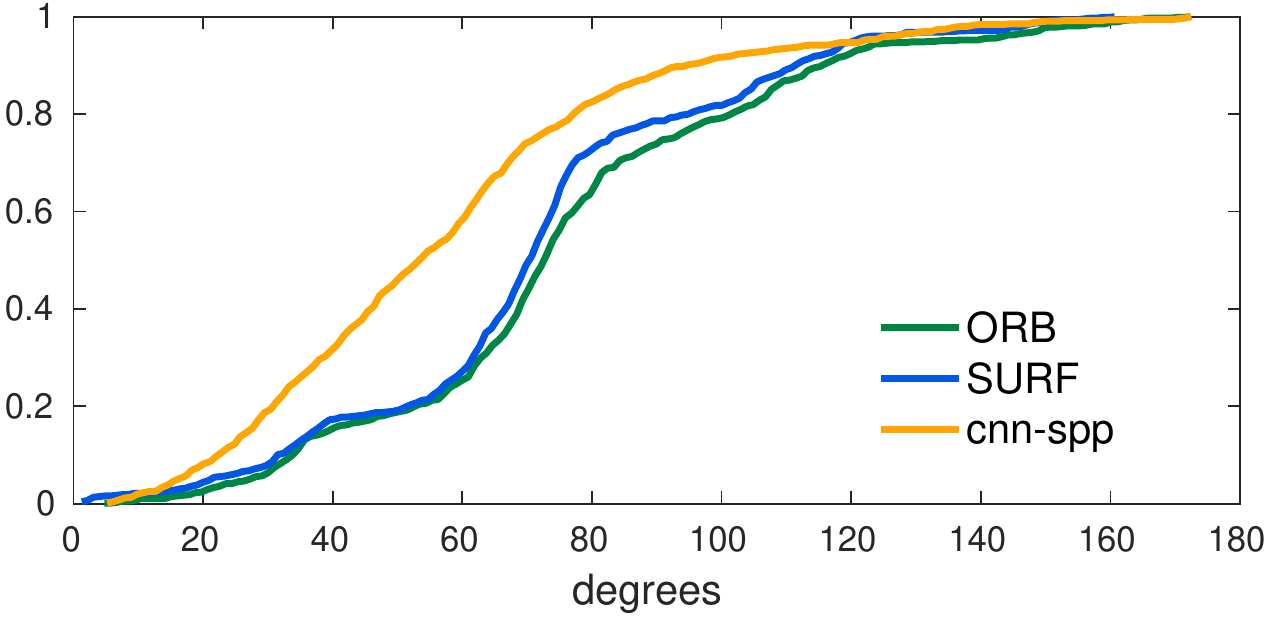}
\includegraphics[width=.45\textwidth]{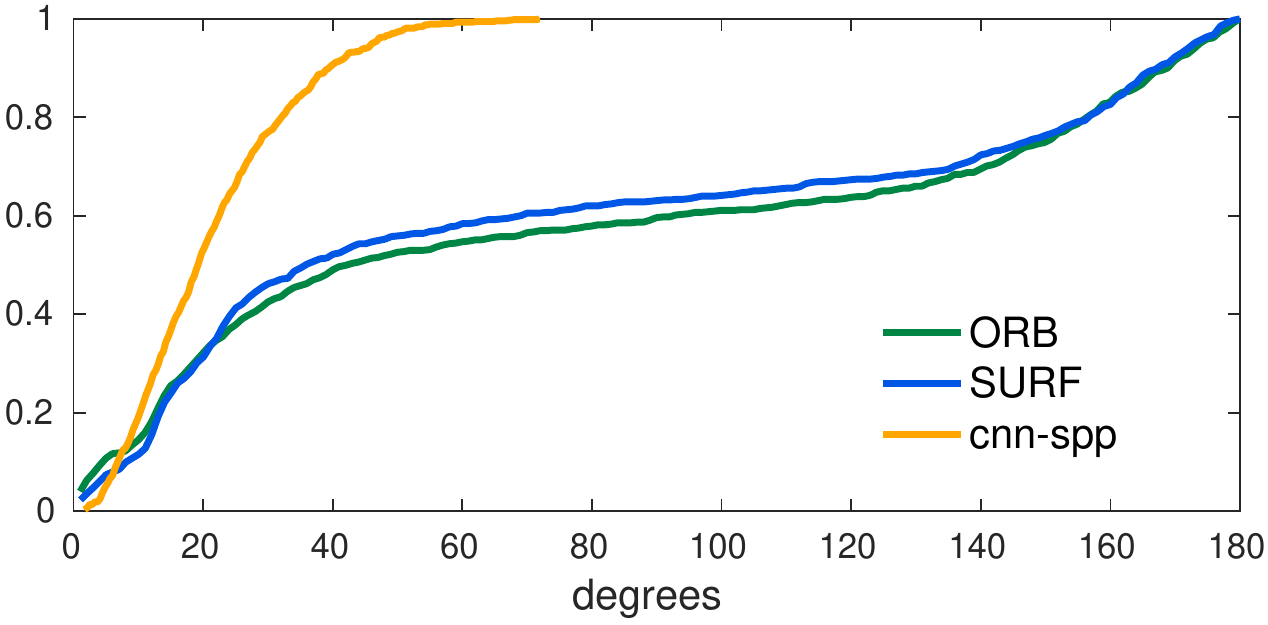}
\includegraphics[width=.45\textwidth]{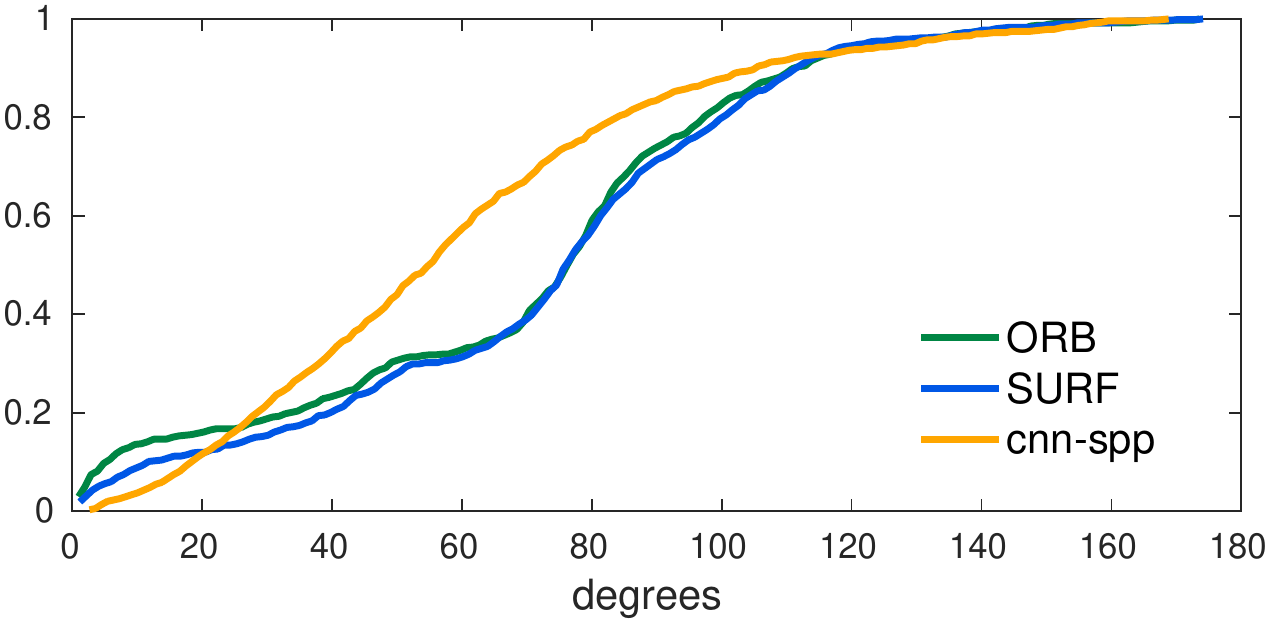}
\caption{Our CNN-based method performs clearly better than conventional local feature based approaches in estimating relative camera orientation (left column) and translation (right column) for the hard cases visualized in Fig.~\ref{subfig:head_exmpl}. Particularly, neither SURF nor ORB are able to localize sufficient amount of inliers for such scenes, and, hence, their performance is quite poor.}\label{subfig:evaluation_on_head_and_rabbit}
\end{subfigure}
\caption{A comparison of traditional point-based methods and the proposed CNN-based approach for estimating relative camera pose. The first row (Fig.~\ref{subfig:performance_all_test_scenes}) shows that in general cnn-spp predicts relative orientation more accurately than SURF or ORB descriptor, but in some cases falls behind in estimating relative translation. However, for cases (Fig.~\ref{subfig:head_exmpl}) where point-based methods are not able to detect enough features, cnn-spp performs significantly better (Fig.~\ref{subfig:evaluation_on_head_and_rabbit}). Furthermore, utilizing high resolution DTU images during training can further improve the results (Fig.~\ref{subfig:performance_all_test_scenes}).}\label{fig:final_results_test_mvs}
\end{figure}

The final results, illustrated in Fig.~\ref{fig:final_results_test_mvs}, show that the fine-tuned model produces more accurate relative camera orientation estimations than the one trained just on Landmarks dataset (red and grey curves respectively). Further, utilizing high resolution images both during training and evaluation leads to the best performance  among the CNN-based methods (yellow curve). On average the proposed method falls slightly behind the feature based approaches in estimating the relative translation. However, based on our inspection it can be said that in certain cases (see for example Fig.~\ref{subfig:head_exmpl}) where the objects are mostly textureless the proposed CNN-based camera pose regressor significantly outperforms traditional feature based approaches (Fig.~\ref{subfig:evaluation_on_head_and_rabbit}).

\vspace{-3mm}
\section{Discussion and Conclusion} \label{sec:conc} 
\vspace{-3mm}
We presented an end-to-end trainable convolutional neural network based approach for estimating the relative pose between two cameras. We evaluated several different network architectures which all were based on the idea of combining two identical network structures with weight sharing, i.e. the Siamese network. We showed that the problem is solvable by this structure and that it is useful to have a larger feature map fed as input to the stack of final fully-connected layers. We further showed that applying spatial pyramid pooling is the key to even more accurate relative pose estimations as it opens the door for larger images that according to our results is one way to improve the accuracy. Overall, our proposal demonstrates very promising results, but there is also some room for further improvement. 

For future directions, one interesting option is to construct a model based on two steps. During the first step, a CNN would produce coarse estimations and then, on the second step, would further refine the final estimate using the preliminary estimations. This kind of a model is reminiscent of the one presented in \cite{demon}. We leave constructing such a model for future work.

%
\bibliographystyle{splncs}
\bibliography{egbib}

\end{document}